\documentclass{article}

\usepackage{microtype}
\usepackage{graphicx}
\usepackage{subfigure}
\usepackage{booktabs} 
\usepackage{float}

\usepackage{hyperref}

\usepackage[accepted]{icml2021}

\icmltitlerunning{Enhancing Visual Realism: Fine-Tuning InstructPix2Pix for Advanced Image Colorization}

\begin{document}

\twocolumn[
\icmltitle{Enhancing Visual Realism: Fine-Tuning InstructPix2Pix for Advanced Image Colorization}

\begin{icmlauthorlist}
\icmlauthor{Zifeng An}{}
\icmlauthor{Zijing Xu}{}
\icmlauthor{Eric Fan}{}
\icmlauthor{Qi Cao}{}
\end{icmlauthorlist}


\vskip 0.3in
]


\begin{abstract}

This paper presents a novel approach to human image colorization by fine-tuning the InstructPix2Pix model, which integrates a language model (GPT-3) with a text-to-image model (Stable Diffusion). Despite the original InstructPix2Pix model's proficiency in editing images based on textual instructions, it exhibits limitations in the focused domain of colorization. To address this, we fine-tuned the model using the IMDB-WIKI dataset, pairing black-and-white images with a diverse set of colorization prompts generated by ChatGPT. This paper contributes by (1) applying fine-tuning techniques to stable diffusion models specifically for colorization tasks, and (2) employing generative models to create varied conditioning prompts. After finetuning, our model outperforms the original InstructPix2Pix model on multiple metrics quantitatively, and we produce more realistically colored images qualitatively. The code for this project is provided on the GitHub Repository \url{https://github.com/AllenAnZifeng/DeepLearning282}.

\end{abstract}

\section{Introduction}
\label{sec:intro}

\begin{figure}[htpb]
    \centering
    \includegraphics[width=0.98\linewidth]{ 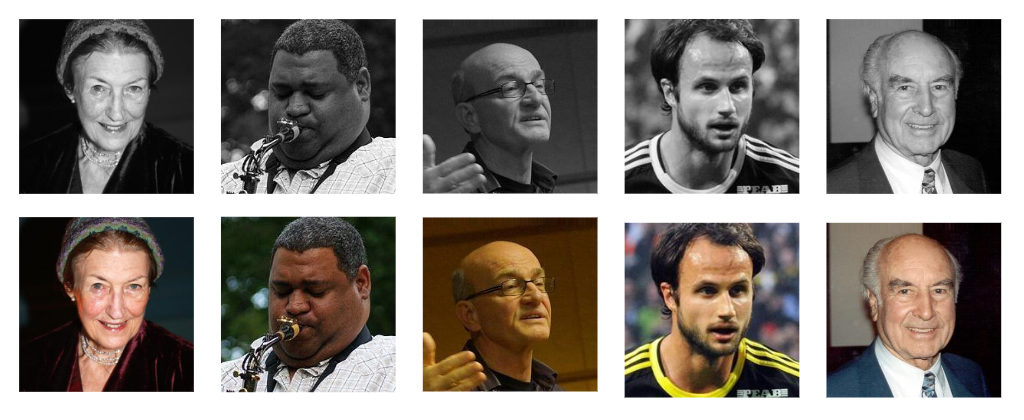}
    \caption{Example of colorization for human faces from the IMDB-WIKI dataset~\cite{Rothe-IJCV-2018}. The task is a pixel-level task that should assign a color to each pixel in the input image. The top row shows the input of the task, a black-and-white image, and the bottom row is the ground truth labels for the input images.}
    \label{fig:colorization_example}
\end{figure}

Image colorization is the task of adding plausible color information to monochromatic images~\cite{vitoria2020chromagan}. Colorization has broad applications, ranging from restoring historical photographs and films to aiding data visualization to enhance the interpretability of medical and satellite imagery. An example of the application of colorization on human faces is shown in Figure~\ref{fig:colorization_example}. The advent of digital technology introduced the concept of automatically colorizing images~\cite{10.1145/2897824.2925974, zhang2016colorful}. However, there are many challenges in this field for computers to understand and interpret the context, texture, and shadows in a monochromatic image so that they can apply realistic colors.

Recent developments in machine learning based computer vision techniques enabled more sophisticated and nuanced approaches to colorization. Many works used Convolutional Neural Networks (CNN) to capture the features in an image~\cite{zhang2016colorful} and used the LAB colorspace for predicting the color of each pixel. Other approaches incorporated both local and global features while training the CNN~\cite{zhang2016colorful, 10.1145/2897824.2925974}. Generative Adversarial Networks (GAN) also proved to be effective in this field. Among GANs, Isola et al. introduced the Pix2Pix framework, a groundbreaking conditional GAN, to generate high-quality, detailed images~\cite{isola2018imagetoimage}.

We decided to finetune a model for colorization based on InstructPix2Pix, an innovative advancement to the original Pix2Pix model through pairing a language model (GPT-3) with a text-to-image model (Stable Diffusion)~\cite{brooks2023instructpix2pix}. It provides a conditional diffusion model such that given an input image and a text instruction, it generates an edited image. The model is trained such that it achieves zero-shot generalization for arbitrary real images and natural human-written instructions. However, the model may have lower performances on a specified transformation, so we decided to finetune the model to complete transformations specifically for colorization.

Our project mainly contributes:
\begin{itemize}
    \item Using finetuning on stable diffusion models for colorization;
    \item Using multiple prompts generated by generative models to diversify conditioning;
\end{itemize}

\section{Methods}
\label{sec:methods}

The original InstructPix2Pix model is designed to edit images based on textual instructions. However, the original model does not perform well when it is only evaluated on the task of colorization. To finetune InstructPix2Pix for colorization, we used the IMDB-WIKI dataset~\cite{Rothe-IJCV-2018} with instructions co-created by ChatGPT. During finetuning, we froze several components unrelated to our task of colorization and focused on finetuning the models related to the image generation.

\subsection{Dataset Preparation}


We utilized a subset of 766 images from the IMDB-WIKI dataset, which is the largest publicly available dataset of facial images for training~\cite{Rothe-IJCV-2018}. We transformed the original colored images into black-and-white images as inputs for the model. The original colored images were used as ground truth to compute loss. Some example images from the dataset are displayed in Figure~\ref{fig:colorization_example}.

Driven in part by the refinement directives articulated in FLAN-V2~\cite{wei2021finetuned}, which underscored the efficacy enhancement associated with the incorporation of Cognitive Tasking (CoT) data, we adopted a methodology that generated 30 synonymous prompts based on the base prompt ``colorize the image''. We used GPT-4 to generate these prompts and combined them with our images to construct our dataset.\footnote{Our dataset can be found on the link \url{https://huggingface.co/datasets/annyorange/colorized_people-dataset}} This approach was designed not only to fortify the robustness of testing procedures but also to optimize overall performance. For the validation dataset, we deliberately adhered to employing solely the prompt ``colorize the image,'' ensuring a consistent basis for meaningful comparisons.

\subsection{Finetuning Approaches}
\begin{figure}[!htpb]
    \centering
    \includegraphics[width=0.98\linewidth]{ 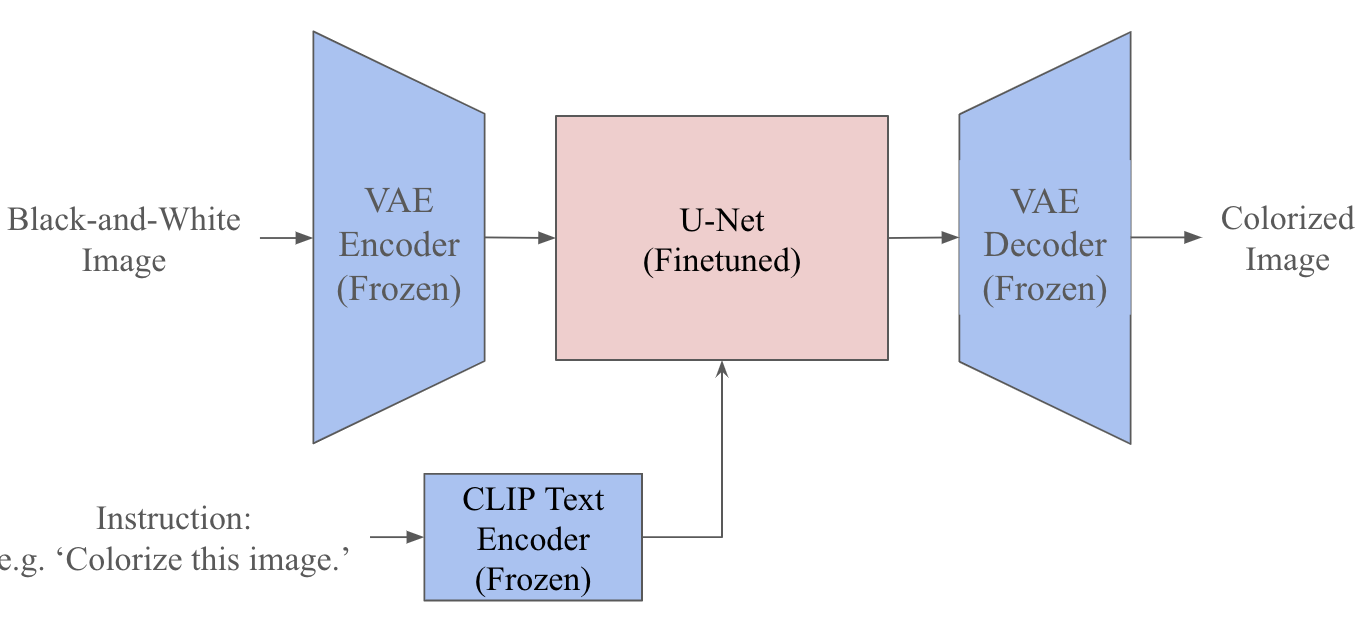}
    \caption{Model architecture of InstructPix2Pix. For our finetuning on the task of colorization, we froze the VAE (encoder and decoder) and the CLIP text encoder. The U-Net is not frozen and is finetuned in the process.}
    \label{fig:arch}
\end{figure}

The InstructPix2Pix architecture is comprised of several key components: the Variational Auto-Encoder (VAE) model (AutoencoderKL) for encoding and decoding images to and from latent representations, the text-encoder from CLIP for encoding the textual instructions, and a conditional U-Net for denoising the encoded image latents~\cite{brooks2023instructpix2pix}. The VAE and CLIP focused on encoding the image and instruction into latent space, which is more general and not directly related to our task of colorization. Therefore, we froze these two models in the InstructPix2Pix pipeline and only finetuned the U-Net, which was in charge of denoising the noisy latents into the desired distribution of colored images. Figure~\ref{fig:arch} shows the architecture of InstructPix2Pix and which models within the pipeline we froze or finetuned. 

When finetuning our model, we employed two distinct loss functions to optimize performance: training loss and validation loss. The training loss is the traditional loss for stable diffusion between the predicted and actual noise. The validation loss is calculated based on a pixel-wise Mean Squared Error (MSE) between the generated images and the original ground truth images, specifically within the LAB color domain. This approach ensures that the colorization process faithfully reproduces accurate and realistic colors by closely mirroring the true color values of the original images.

During this process, we adjusted various hyperparameters, namely the learning rate, batch size, and prompt size. The results of varying these hyperparameters are shown in Section~\ref{sec:hyperparam}.

\begin{figure}[htpb]
    \centering
    \includegraphics[width=0.95\linewidth]{ 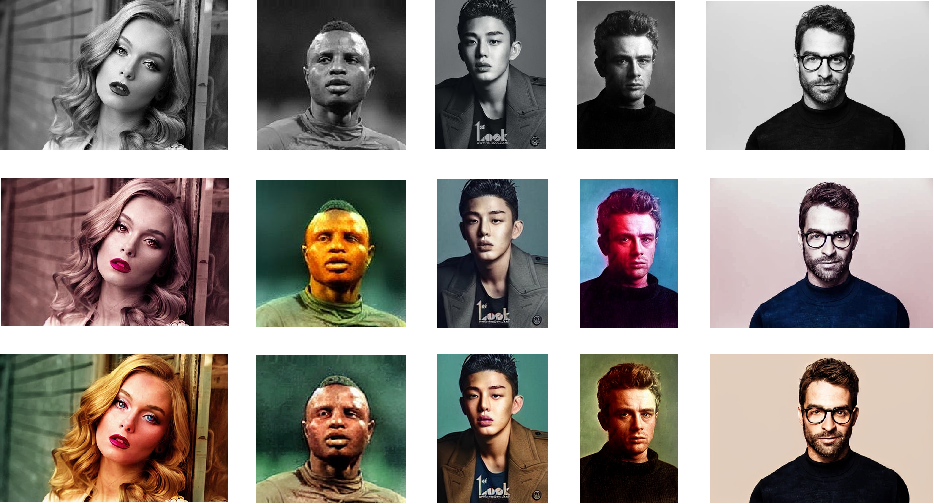}
    \caption{Visual comparison of black-and-white input images (the first row), images colorized by original InstructPix2pix \cite{brooks2023instructpix2pix} (the second row) and colorized by our finetuned model (the third row). Our method is able to accomplish visually pleasant colorization and our result is significantly better than the original version.}
    \label{fig:image_results}
\end{figure}

\section{Results}
\label{sec:results}

\begin{figure}[htpb]
    \centering
    \includegraphics[width=0.95\linewidth]{ 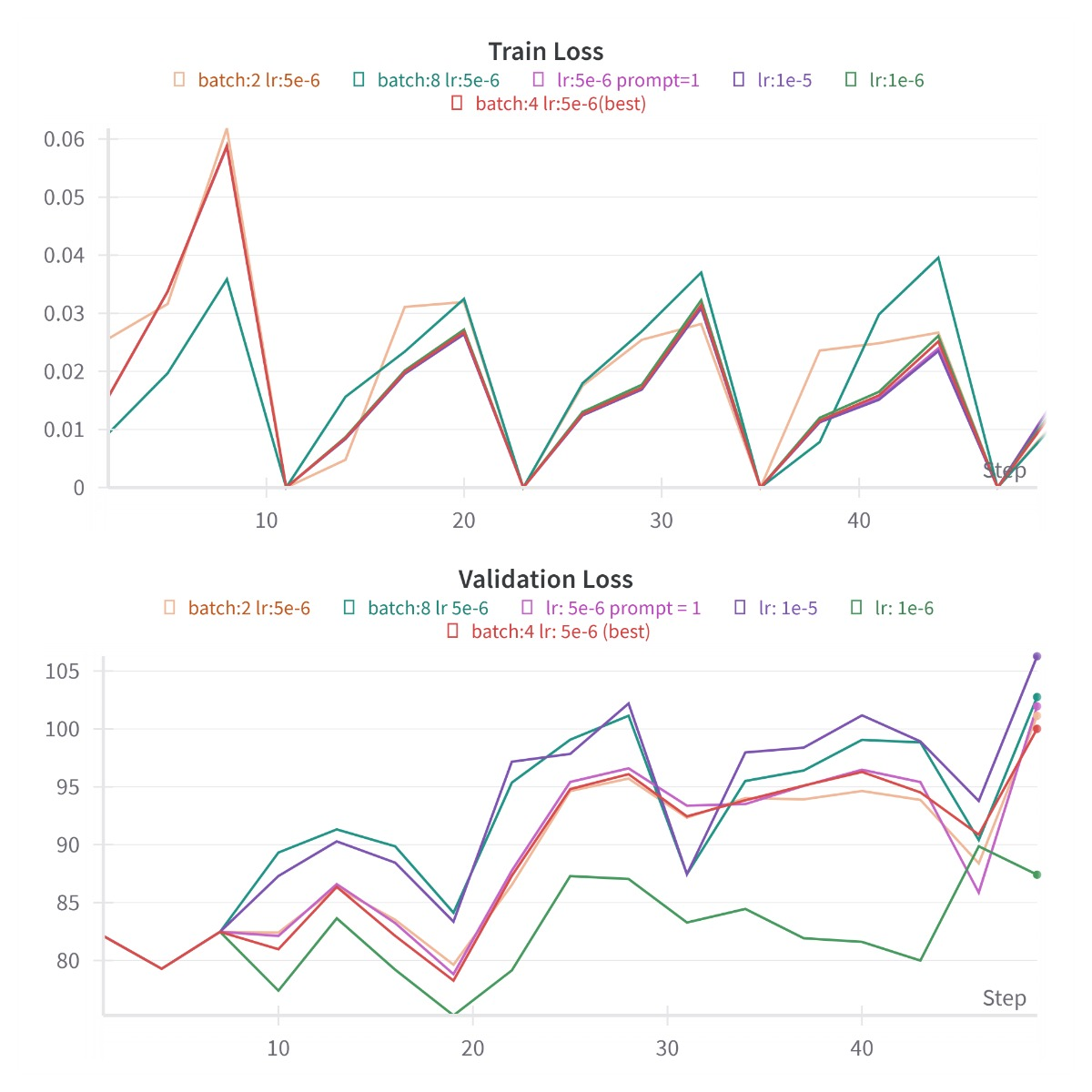}
    \caption{The training loss and the validation loss for our fine-tuned model.}
    \label{fig:losses}
\end{figure}

We conducted fine-tuning for 50 updates, and Figure~\ref{fig:image_results} offers a visual comparison between our best fine-tuned model and the original InstructPix2Pix model. Figure~\ref{fig:losses} displays the training and validation losses for various learning rates, batch sizes, and prompt sizes.

In Figure~\ref{fig:losses}, the training loss, represented by the Mean Squared Error (MSE) loss between the predicted noise and the true noise, exhibits oscillations. This behavior is expected as the model learns to minimize the discrepancy between generated and actual noise. The peaks in the training loss gradually decrease, indicating the model's convergence to a better fit during fine-tuning.

\begin{figure}[htpb]
    \centering
    \includegraphics[width=0.98\linewidth]{ 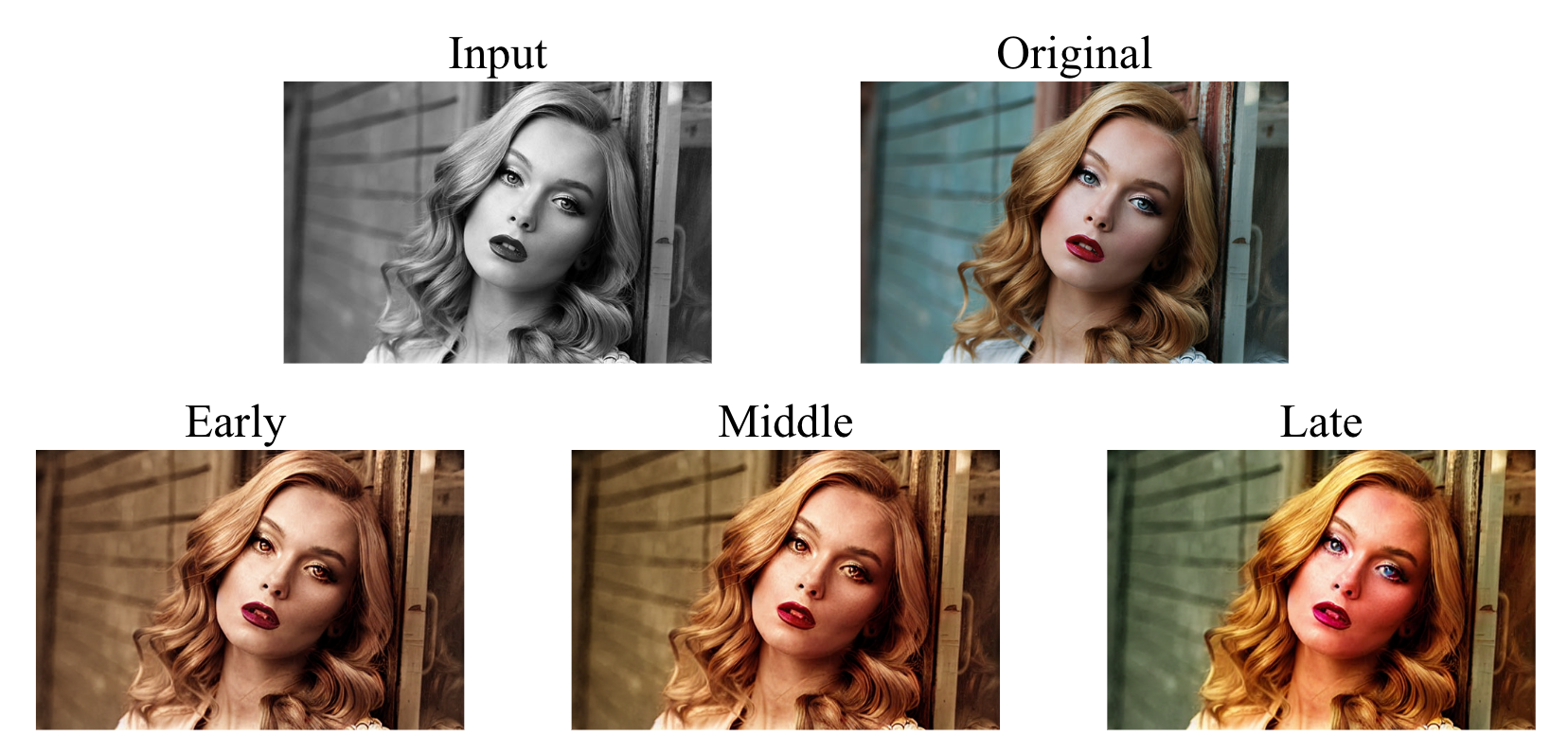}
    \caption{Comparison of the colorization abilities on a sample image at different training steps of finetuning. The first row shows the input image and the original colorized image, and the second row shows the early, middle, and late stages of finetuning, from left to right respectively.}
    \label{fig:stage_example}
\end{figure}

Conversely, the validation loss, measuring the MSE loss between the predicted colorized image and the original image, shows an increasing trend with the number of training steps. This phenomenon is attributed to the limitation of MSE in perfectly reflecting the absolute quality of colorization. For instance, in colorization tasks, different colors can be valid for the same black-and-white photo under similar lighting conditions. Additionally, the choice of step length during training is a parameter to consider, as evidenced by the different colorization stages in Figure~\ref{fig:stage_example}.

To further interpret the training process, we captured images at early, middle, and late stages of fine-tuning, as depicted in Figure~\ref{fig:stage_example}. This visualization underscores the evolution of colorization quality throughout training, providing insights into how the model's performance improves over time.

\subsection{Quantitative Comparison}

\begin{table}[h]
    \centering
    \caption{\textbf{Quantitative comparison} on IMDB-WIKI datasets. Upward arrows indicate that a higher score denotes good image quality.}
    \resizebox{\linewidth}{!}{
        \begin{tabular}{lccc}
            \toprule
            \multicolumn{4}{c}{\textbf{Dataset  IMDB-WIKI}} \\
            \midrule
            \textbf{Metric} & \textbf{PSNR \textuparrow} & \textbf{SSIM \textuparrow} & \textbf{MAE \textdownarrow} \\
            \midrule
            InstructPix2pix & 19.7804 & 0.4301 & 22.1093 \\
            Fine-Tuned Model & \textbf{19.9019} & \textbf{0.5348} & \textbf{21.3045} \\
            \bottomrule
        \end{tabular}
    }
    \label{tab:imdb-wiki-comparison}
\end{table}

For a comprehensive assessment, we utilized common image quality metrics: Peak Signal-to-Noise Ratio (PSNR), Structural Similarity Index (SSIM)~\cite{wang2004image}, and Mean Absolute Error (MAE). These metrics are widely used in prior studies ~\cite{xu2023pik,he2018deep,su2020instance}. PSNR measures fidelity, with higher values indicating improved image quality; SSIM evaluates structural similarity, where higher scores denote more consistent structures, and MAE quantifies the average absolute pixel-wise difference, reflecting overall image similarity. The quantitative results of these metrics on the IMDB-WIKI dataset are illustrated in Table~\ref{tab:imdb-wiki-comparison}. Our approach outperforms the original InstructPix2Pix models in all of the metrics (PSNR, SSIM, and MAE).

\subsection{Hyperparameter Tuning}
\label{sec:hyperparam}
\begin{figure}[htpb]
    \centering
    \includegraphics[width=0.98\linewidth]{ 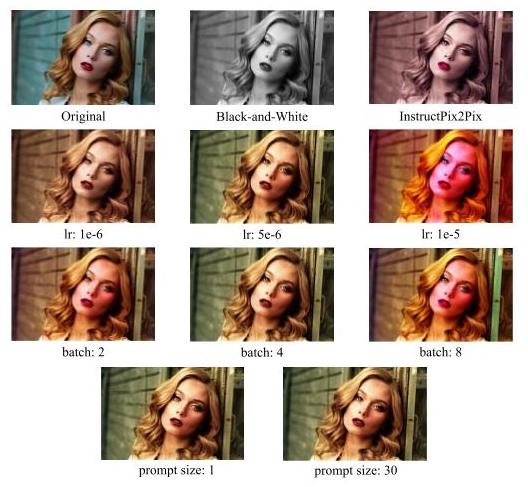}
    \caption{This figure shows the results of this sample input image (black-and-white) to our model with different hyperparameters.}
    \label{fig:hyperparam}
\end{figure}

To achieve our best results, we compared several hyperparameters in our model, namely the learning rate, batch size, and the number of generated prompts we used. The comparison of the resultant images is shown in Figure~\ref{fig:hyperparam}. In default, we used learning rate $5 \times 10^{-6}$, batch size 4, and 30 generated prompts. We change each hyperparameter separately for each section. 

\paragraph{Learning Rate} We set the learning rate to $1 \times 10^{-5}$, $5 \times 10^{-6}$, $1 \times 10^{-6}$. As shown in the second row of Figure~\ref{fig:hyperparam}, the lower learning rate produces an image that is dull and is very close to InstructPix2Pix without much changes, while the higher learning rate produces an image that has very high contrast with vibrant colors. We found the balanced learning rate at $5 \times 10^{-6}$ with the best results.

\paragraph{Batch Size} We used different batch sizes ranging from 2, 4, and 8. The facial colorization looks better in smaller batch sizes, but larger batch sizes provide some more coloring to the background. However, larger batch sizes update the model more given the same number of training steps, so they might overfit the model and cause the original model to start forgetting.

\paragraph{Number of Prompts} We tested our model using only a single prompt versus 30 generated prompts. In both cases, the prompts used for training are different from the prompt ``colorize the image'' used in validation. We intended to vary the number of prompts to see if the model will perform worse if it was not well trained on multiple similar prompts and thus fails to generalize to an unseen prompt. However, from Figure~\ref{fig:hyperparam} there is not much difference between using 1 and 30 prompts. We believe the CLIP text encoder will project similar prompts to similar embeddings, resulting in low differences between the training and validation prompts. One potential approach to improve the prompting mechanism is to introduce soft prompting, thus eliminating the need for hard-coded prompts~\cite{DBLP:journals/corr/abs-2104-08691}. 

\section{Conclusion and Future Work}
\label{sec:conclusion}

In this paper, we presented an innovative approach to image colorization by fine-tuning the InstructPix2Pix model. Our method focused on selectively freezing components of the InstructPix2Pix model and fine-tuning the U-Net for image latent denoising. The results, both qualitative and quantitative, demonstrate a significant enhancement in the model's ability to realistically colorize images. From a qualitative perspective, the colorization results of our model stand out in their visual perception. The colors in the images generated by our model strike a harmonious balance -- they are neither too dull and close to grayscale nor too sharp and overly vibrant. Quantitatively, our model outperforms the original InstructPix2Pix in Peak Signal-to-Noise Ratio (PSNR), Structural Similarity Index (SSIM), and Mean Absolute Error (MAE). Inspired by FLAN-V2, we employed GPT-4 to generate a diverse array of 30 prompts to ensure diverse prompt conditioning in model training.

The current method has areas that can be improved. A primary concern is the inconsistency in the loss metrics: the training loss is based on noise prediction, whereas the validation loss relies on pixel-wise MSE in the LAB color space. This misalignment could potentially limit the model's effectiveness in generalizing to real-world colorization tasks. To enhance this, we propose combining these two loss functions by adding learnable weights, allowing for joint optimization during the training phase. This approach would enable the model to not only accurately predict noise but also improve color fidelity as reflected in the LAB MSE. Such a unified training process is likely to result in more consistent improvements, where enhancements in training loss directly translate to better validation performance.

Further exploration into the effects of various hyperparameters on model performance is also essential. While our study experimented with different learning rates, batch sizes, and prompt quantities, a more in-depth investigation could provide deeper insights into optimal settings for these parameters. This could involve extensive experiments with varying prompt structures or the exploration of soft prompting techniques, offering a more nuanced approach to training.

\bibliography{main}

\begin{thebibliography}{12}
\providecommand{\natexlab}[1]{#1}
\providecommand{\url}[1]{\texttt{#1}}
\expandafter\ifx\csname urlstyle\endcsname\relax
  \providecommand{\doi}[1]{doi: #1}\else
  \providecommand{\doi}{doi: \begingroup \urlstyle{rm}\Url}\fi

\bibitem[Brooks et~al.(2023)Brooks, Holynski, and
  Efros]{brooks2023instructpix2pix}
Brooks, T., Holynski, A., and Efros, A.~A.
\newblock Instructpix2pix: Learning to follow image editing instructions, 2023.

\bibitem[He et~al.(2018)He, Chen, Liao, Sander, and Yuan]{he2018deep}
He, M., Chen, D., Liao, J., Sander, P.~V., and Yuan, L.
\newblock Deep exemplar-based colorization.
\newblock \emph{ACM Transactions on Graphics (TOG)}, 37\penalty0 (4):\penalty0
  1--16, 2018.

\bibitem[Iizuka et~al.(2016)Iizuka, Simo-Serra, and
  Ishikawa]{10.1145/2897824.2925974}
Iizuka, S., Simo-Serra, E., and Ishikawa, H.
\newblock Let there be color! joint end-to-end learning of global and local
  image priors for automatic image colorization with simultaneous
  classification.
\newblock \emph{ACM Trans. Graph.}, 35\penalty0 (4), jul 2016.
\newblock ISSN 0730-0301.
\newblock \doi{10.1145/2897824.2925974}.
\newblock URL \url{https://doi.org/10.1145/2897824.2925974}.

\bibitem[Isola et~al.(2018)Isola, Zhu, Zhou, and Efros]{isola2018imagetoimage}
Isola, P., Zhu, J.-Y., Zhou, T., and Efros, A.~A.
\newblock Image-to-image translation with conditional adversarial networks,
  2018.

\bibitem[Lester et~al.(2021)Lester, Al{-}Rfou, and
  Constant]{DBLP:journals/corr/abs-2104-08691}
Lester, B., Al{-}Rfou, R., and Constant, N.
\newblock The power of scale for parameter-efficient prompt tuning.
\newblock \emph{CoRR}, abs/2104.08691, 2021.
\newblock URL \url{https://arxiv.org/abs/2104.08691}.

\bibitem[Rothe et~al.(2018)Rothe, Timofte, and Gool]{Rothe-IJCV-2018}
Rothe, R., Timofte, R., and Gool, L.~V.
\newblock Deep expectation of real and apparent age from a single image without
  facial landmarks.
\newblock \emph{International Journal of Computer Vision}, 126\penalty0
  (2-4):\penalty0 144--157, 2018.

\bibitem[Su et~al.(2020)Su, Chu, and Huang]{su2020instance}
Su, J.-W., Chu, H.-K., and Huang, J.-B.
\newblock Instance-aware image colorization.
\newblock In \emph{Proceedings of the IEEE/CVF Conference on Computer Vision
  and Pattern Recognition}, pp.\  7968--7977, 2020.

\bibitem[Vitoria et~al.(2020)Vitoria, Raad, and
  Ballester]{vitoria2020chromagan}
Vitoria, P., Raad, L., and Ballester, C.
\newblock Chromagan: Adversarial picture colorization with semantic class
  distribution, 2020.

\bibitem[Wang et~al.(2004)Wang, Bovik, Sheikh, and Simoncelli]{wang2004image}
Wang, Z., Bovik, A.~C., Sheikh, H.~R., and Simoncelli, E.~P.
\newblock Image quality assessment: from error visibility to structural
  similarity.
\newblock \emph{IEEE transactions on image processing}, 13\penalty0
  (4):\penalty0 600--612, 2004.

\bibitem[Wei et~al.(2021)Wei, Bosma, Zhao, Guu, Yu, Lester, Du, Dai, and
  Le]{wei2021finetuned}
Wei, J., Bosma, M., Zhao, V.~Y., Guu, K., Yu, A.~W., Lester, B., Du, N., Dai,
  A.~M., and Le, Q.~V.
\newblock Finetuned language models are zero-shot learners.
\newblock \emph{arXiv preprint arXiv:2109.01652}, 2021.

\bibitem[Xu et~al.(2023)Xu, Tu, Du, Dong, Li, Meng, Ma, Bovik, and
  Yu]{xu2023pik}
Xu, R., Tu, Z., Du, Y., Dong, X., Li, J., Meng, Z., Ma, J., Bovik, A., and Yu,
  H.
\newblock Pik-fix: Restoring and colorizing old photos.
\newblock In \emph{Proceedings of the IEEE/CVF Winter Conference on
  Applications of Computer Vision}, pp.\  1724--1734, 2023.

\bibitem[Zhang et~al.(2016)Zhang, Isola, and Efros]{zhang2016colorful}
Zhang, R., Isola, P., and Efros, A.~A.
\newblock Colorful image colorization, 2016.

\end{thebibliography}
\bibliographystyle{icml2021}

\clearpage

\end{document}